\documentclass[]{spie}  

\usepackage{amsmath,amsfonts,amssymb}
\usepackage{graphicx}
\usepackage[colorlinks=true, allcolors=blue]{hyperref}
\usepackage[font=footnotesize]{subfig}
\usepackage{threeparttable}

\title{A Novel Scene Text Detection Algorithm Based On Convolutional Neural Network}

\author{Xiaohang Ren, Kai Chen, Jun Sun}
\affil{Institute of Image Communication and Network Engineering, Shanghai Jiao Tong University, Shanghai, China.}

\authorinfo{Further author information: (Send correspondence to Kai Chen)\\E-mail: [xiaomu, kchen, junsun]@sjtu.edu.cn}

\begin{document}
\maketitle

\begin{abstract}
Candidate text region extraction plays a critical role in convolutional neural network (CNN) based text detection from natural images. In this paper, we propose a CNN based scene text detection algorithm with a new text region extractor. The so called candidate text region extractor I-MSER is based on Maximally Stable Extremal Region (MSER), which can improve the independency and completeness of the extracted candidate text regions. Design of I-MSER is motivated by the observation that text MSERs have high similarity and are close to each other. The independency of candidate text regions obtained by I-MSER is guaranteed by selecting the most representative regions from a MSER tree which is generated according to the spatial overlapping relationship among the MSERs. A multi-layer CNN model is trained to score the confidence value of the extracted regions extracted by the I-MSER for text detection. The new text detection algorithm based on I-MSER is evaluated with wide-used ICDAR 2011 and 2013 datasets and shows improved detection performance compared to the existing algorithms.
\end{abstract}


\section{INTRODUCTION}
\label{sec:intro}  

Text detection from scene images is a crucial technology in large numbers of computer vision applications, such as image classification, scene recognition and automatic navigation in urban environments. In document domain, OCR systems is very mature and have achieved very good performance in detecting texts from scan documents. However, text detection from scene images is much more challenging due to the variations of texts in font, size and style, complex backgrounds, noise, unconfirmed lighting conditions (like using flash lamps), and geometric distortions. To quantify and track the progress of text location in natural images, several competitions, including the four ICDAR Text Location Competitions in 2011 and 2013~\cite{ICDAR2011,ICDAR2013} have been held in recent years. Recently many text detection algorithms have been reported in the literatures~\cite{Epshtein2010,Shivakumara2011,Neumann2012,Tsai2012}.Most of them use a candidate text region extractor and one or several advanced features such as HOG or SIFT to detect text regions by a classifier or some heuristic rules. MSER, an efficient key-region detector proposed by Matas et al.~\cite{Matas2004}, is the most widely used candidate text region extractor and achieves great success in scene text detection. However, it has been shown in the literature that a considerable number of candidate text regions extracted by MSER contain both text and non-text parts, which makes classification task very difficult. A modified MSER detector is presented in~\cite{Chen2011}, in which the MSER is edge-enhanced for improving the independency of the candidate text regions. But in the meanwhile, the number of non-text regions is significantly increased, which also increases the difficulty of classification.

Recently a growing amount of research on visual recognition turned to use deep learning tools, which can extract more accurate image features in representing the particular images than the hand-crafted features. A convolutional neural network (CNN) based text detection algorithm is presented in~\cite{wang2012}. In order to extract all the text regions with suitable scales for CNN, the multi-scale sliding window method extracts massive image patches most of which are non-text regions. The large number of non-text regions increases the computational complexity of the CNN model and limits the performance of the text detection algorithm.

In this paper, we make two major contributions. First, we develop a novel modified MSER extractor, I-MSER, which is robust against noise, contrast variations, and complex backgrounds. I-MSER is designed based on the many overlapping regions extracted by MSER which are caused by the image blur in text regions. The statistics of text and background regions are significantly different in appearance, which motivates us to design the I-MSER detector. The regions extracted by I-MSER are the most representative regions in the MSER trees. In I-MSER, every pixel can only belong to one region, which guarantees the independency of extracted regions. And the completeness is guaranteed by merging similar neighbor regions. Second, we optimize main parameter of I-MSER through statistical region analyze to minimize the number of extracted regions with sufficient text regions. Taking the advantages of optimized I-MSER, the multi-layer CNN model can identify the text regions with much higher accuracy and less computational complexity.

\section{THE TEXT DETECTION ALGORITHM}



\subsection{The Motivation of Isolated MSER}
\label{sec:title}

\begin{figure}[!tbp]
  \centering
    \subfloat[]{
        \includegraphics[width=0.22\linewidth]{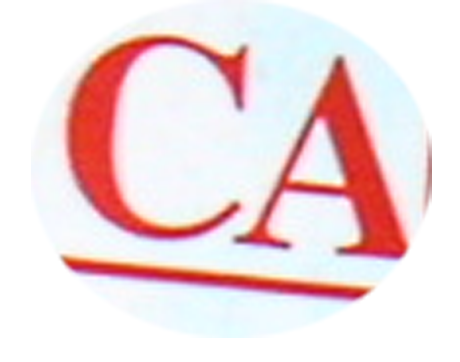}
        \centering
        \label{textdiff:a}
    }
    \subfloat[]{
        \centering
        \includegraphics[width=0.22\linewidth]{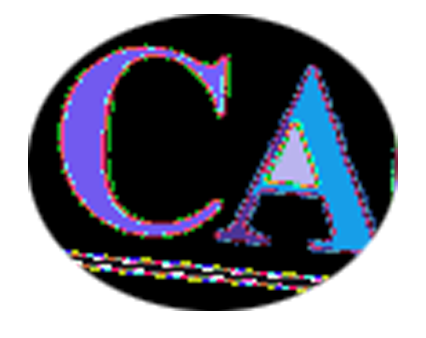}
        \label{textdiff:b}
    }
    \subfloat[]{
        \centering
        \includegraphics[width=0.22\linewidth]{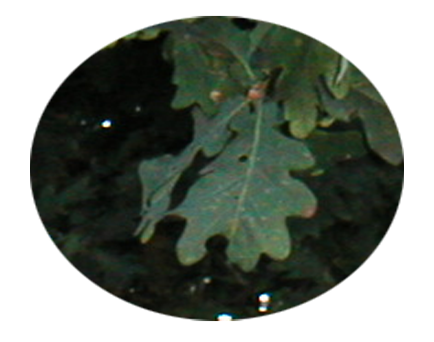}

        \label{textdiff:c}
    }
    \subfloat[]{
        \centering
        \includegraphics[width=0.22\linewidth]{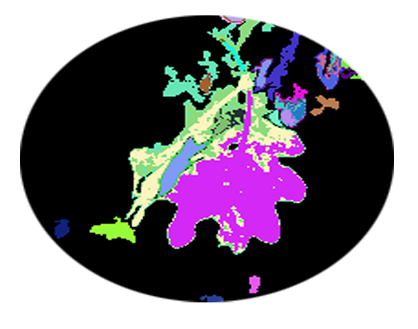}
        \label{textdiff:d}
    }


  \caption{The text and background regions difference.(a) Typical text region. (b) Text MSER(additional pixels are in different color). (c) Typical background region. (d) Background MSER.
  }
  \label{textdiff}
\end{figure}

The widely used implementation of MSER~\cite{Nister2008} has additional parameters to control the output regions. One of the most important parameters is $minDiversity$, which affects the number of output regions by controlling the merging of overlapping MSER regions. With low $minDiversity$, MSER extracts a large number of overlapping regions, which is regarded as a performance degeneration against blur. However, as texts are normally with outside sharp edges and backgrounds are with inside weak edges, the overlapping regions can be important information in dividing them. As shown in Fig.\ref{textdiff}(a), texts have outside sharp edges which may be blurred by compression algorithm like JPEG. Therefore, there are many highly similar MSERs in the text regions, as depicted in Fig.\ref{textdiff}(b). Those outer MSERs are just one or two pixel wider then the inner MSERs. The background regions, especially complex backgrounds like plant regions, have many weak edges inside, as shown in Fig.\ref{textdiff}(c). Thus the outer regions and inner regions have low similarity, Fig.\ref{textdiff}(d).


\subsection{The Isolated MSER}

I-MSER is a modified MSER detector which outputs non-overlapped MSERs in images. I-MSER extracts the most stable ERs which are probably isolated objects and has good performance in detecting text regions because text regions have significant difference from background regions. 

We generate component trees as done in~\cite{Nister2008}, then choose a small $\Delta$ to get a large number of MSERs and classify them into groups by spatial overlapping relationship. Each group generates a MSER tree and each tree node is a MSER. For a tree $T_n$, it contains a top region denoted by $R_t$ and some root regions denoted by $R_r$. Each non-top region has at least one up region $R_n^u$, and the smallest $R_n^u$ is regarded as the father region $R_n^f$ of region $R_n$. Then we merge $R_n$ and $R_n^f$ if they fulfill the following condition:
\begin{equation}
\frac{|R_n^f|-|R_n|}{|R_n|} \geq \gamma.
\end{equation}
where $\gamma$ is the stability parameter. If they do not fulfill (1), set $R_n$ as an I-MSER, then cut out all the pixels in $R_n$ from every $R_n^u$.
\begin{figure}[!tbp]
  \centering

  \subfloat[]{
    \centering
    \includegraphics[width=0.48\linewidth]{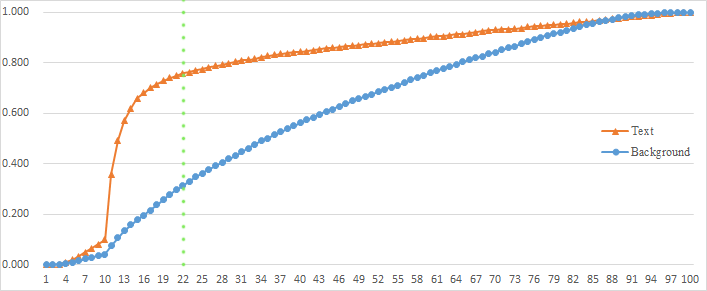}
    \label{cpd}
  }
  \subfloat[]{
    \centering
    \includegraphics[width=0.48\linewidth]{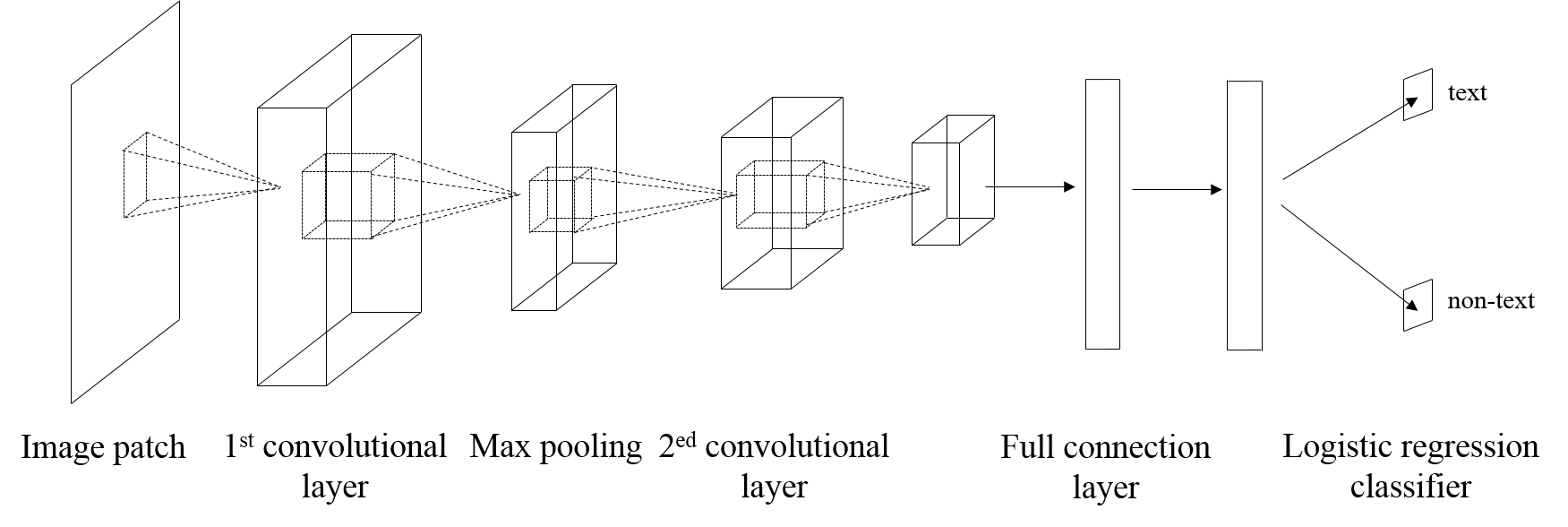}
    \label{cnn}
  }
  \caption{(a). The cumulative proportional distribution(x-axis is the difference ratio and y-axis is the cumulative proportion)
    (b).The CNN model of the algorithm}
    \label{cc}
\end{figure}

To optimize the parameter $\gamma$, the cumulative proportional distributions of region pixel difference in text and background regions are computed. An example of the computation results is shown in Fig.\ref{cc}(a). The dotted line presents results with a certain $\gamma$ value, and the points of intersection between the dotted line and the cumulative probably curves present the cumulative proportions $P_{t,\gamma}$ and $P_{b,\gamma}$ of corresponding to the text and background regions with the $\gamma$ value. The optimal value $\gamma_o$ is calculated by:
\begin{equation}
\gamma_o=max(P_{t,\gamma}-P_{b,\gamma}).
\end{equation}


\subsection{Convolutional Neural Network}

After the I-MSER extraction, the extracted regions are simply filtered and merged to remove the obvious non-text regions and merge the regions describing the same character. We filter the regions by limiting the number of their inner holes, because a character will not contain a large number of holes. Then the regions are merged into a complete candidate text region if they are vertically connected and share similarity in width, size and color. After filtering and merging, the candidate text regions are input to a CNN model for confidence scoring. Finally the score map of the candidate text regions is analysed by a text line formation method, which is based on the basics of the regions such as height, vertical position, spaces between the nearest region and color, to form the detected text lines.

The CNN model in our text detection algorithm has 6 layers including two convolutional layers, two down-sampling layers, and two full-connecting layer. The input image size is fixed to $32\times 32$ by considering the human identifiable text size. The two convolutional layers have 64 and 96 filters with $6\times 6$ and $4\times 4$ kernel sizes, respectively. The down-sampling layers are average-pooling with poolsize of $3\times3$ and $2\times 2$, respectively. The output feature maps are input to two full-connecting layers in size of 200 with drop out. The final output of the CNN is input to a simple softmax classifier. After the CNN model is trained, the soft-max classifier will assign confidence value for every input patches. The CNN model is shown in Fig.\ref{cc}(b).

\section{EXPERIMENTS}
\label{sec:sections}

Our I-MSER text detection algorithm is evaluated with both ICDAR 2011 dataset and ICDAR 2013 dataset, which are the most primary text detection datasets. With the ICDAR 2011 dataset, our algorithm is compared with existing approaches using ER~\cite{Neumann2012}, MSER~\cite{Tsai2012}, SWT~\cite{Epshtein2010}, and the top method in the ICDAR 2011 robust reading competition~\cite{ICDAR2011}. The work in~\cite{Huang2014} is another CNN based text detection algorithm is not compared in the experiment because the evaluation method in~\cite{Huang2014} is not as same as the one of ICDAR 2011. With the ICDAR 2013 dataset, our algorithm is compared with the top three methods in the ICDAR 2013 robust reading competition~\cite{ICDAR2013}. 

\begin{figure}[!tbp]
  \centering
    \subfloat[]{
        \includegraphics[width=0.22\linewidth]{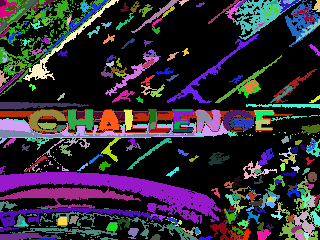}
        \centering
        \label{exres:a}
    }
    \subfloat[]{
        \includegraphics[width=0.22\linewidth]{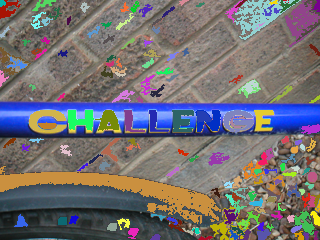}
        \label{exres:b}
    }
    \subfloat[]{
        \centering
        \includegraphics[width=0.26\linewidth]{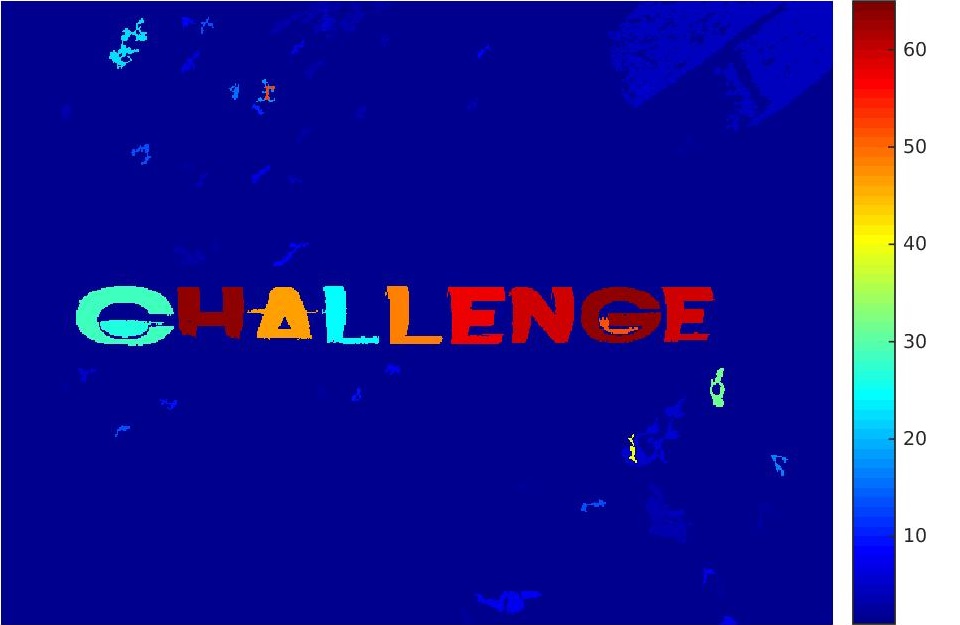}

        \label{exres:c}
    }
    \subfloat[]{
        \centering
        \includegraphics[width=0.22\linewidth]{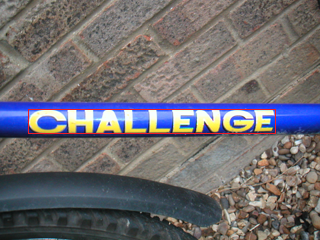}
        \label{exres:d}
    }

  \caption{Intermediate steps in text detection.(a) I-MSER extracted regions
  . (b) The regions after filtering and merging (c) The confidence map generated by CNN
  . (d) The text line extracted by our algorithm
  .}
  \label{exres}
\end{figure}

The intermediate results of our algorithm are shown in Fig.\ref{exres}. Fig.\ref{exres}(a) presents the output of I-MSER, which contains various isolated regions. The I-MSERs exist mostly in background regions and some in text regions. Note that in the first character ``E'', one region is located in the upper part and the other is located in the lower part. And in Fig.\ref{exres}(b), note that the two regions have been merged into one region which presents the character ``E'' individually. We can also easily figure out that a large number of regions in background have been filtered out. Fig.\ref{exres}(c) presents the confidence map of the regions is generated by the CNN model. All the text regions and few background regions gain high confidence value, which shown that the confidence value assigned by the CNN model is highly credible. The final result is shown in Fig.\ref{exres}(d). The text regions are grouped into a text line by the text line formation method while the high confidence background regions are not grouped into text lines because they do not satisfy the formation rules. The boundary is established by the up-left and down-right pixels of the text line.

\begin{table}[!h]
\centering
\caption{\textbf{Evaluation with different $\gamma$ values}}\label{rres}
\begin{threeparttable}
    \begin{tabular}{|c|c|c|c|}
      \hline
      ICDAR 2011 Dataset & $p$ & $r$ & $f$ \\
      \hline
      $\gamma_o-5\%$ & 0.88 & 0.58 & 0.70 \\
      \hline
      $\gamma_o-2\%$ & 0.84 & 0.67 & 0.75 \\
      \hline
      $\gamma_o\%$ & 0.82 & 0.71 & 0.76 \\
      \hline
      $\gamma_o+2\%$ & 0.79 & 0.72 & 0.75 \\
      \hline
      $\gamma_o+5\%$ & 0.77 & 0.73 & 0.74 \\
      \hline
    \end{tabular}
\end{threeparttable}

\end{table}

Table \ref{rres} summarizes the evaluation results of our algorithm in different $\gamma$ (15\%)values. The optimal value $\gamma_o$ is computed by formula (2) with the training set of ICDAR 2011 competition database ~\cite{ICDAR2011}. $\gamma_o$ is chosen for standard performance evaluation, and $\gamma$ with the values of $\gamma_o\pm2\%$ and $\gamma_o\pm5\%$ compare with it. We can note that recall is directly proportional to $\gamma$, while precision is inversely proportional to $\gamma$. Recall with $\gamma_o-5\%$ drops significantly because the extracted text regions are much less than with $\gamma_o$. While with small range above $\gamma_o$, precision drops slowly because the number of extracted text regions increases nearly synchronously with the number of extracted non-text regions and our algorithm has good performance amongst those false.

\begin{table}[!h]
    \begin{minipage}[t]{0.5\linewidth}
        \centering
        \caption{\textbf{Evaluation with ICDAR 2011 dataset}}\label{ICDARres1}
        \begin{threeparttable}
            \begin{tabular}{|c|c|c|c|}
              \hline
              ICDAR 2011 Dataset & $P$ & $R$ & $F$ \\
              \hline
              Our algorithm & 0.82 & \textbf{0.71} & \textbf{0.76} \\
              \hline
              Kim's algorithm~\cite{ICDAR2011} & \textbf{0.83} & 0.62 & 0.71 \\
              \hline
              Tsai's algorithm~\cite{Tsai2012} & 0.73 & 0.66 & 0.69 \\
              \hline
              Neumann's algorithm~\cite{Neumann2012} & 0.73 & 0.65 & 0.69 \\
              \hline
              Epshtein's algorithm~\cite{Epshtein2010} & 0.73 & 0.60 & 0.66 \\
              \hline
            \end{tabular}
        \end{threeparttable}
    \end{minipage}
    \begin{minipage}[t]{0.5\linewidth}
        \centering
        \caption{\textbf{Evaluation with ICDAR 2013 dataset}}\label{ICDARres3}
        \begin{threeparttable}
            \begin{tabular}{|c|c|c|c|}
              \hline
              ICDAR 2013 Dataset & $P$ & $R$ & $F$ \\
              \hline
              Our algorithm & 0.83 & \textbf{0.71} & \textbf{0.77} \\
              \hline
              Yin's algorithm~\cite{ICDAR2013} & \textbf{0.88} & 0.66 & 0.76 \\
              \hline
              Neumann's algorithm~\cite{neumann2013} & 0.88 & 0.65 & 0.74 \\
              \hline
              Bai's algorithm~\cite{bai2013} & 0.79 & 0.68 & 0.73 \\
              \hline
            \end{tabular}
        \end{threeparttable}
    \end{minipage}
\end{table}

Table 2 summarizes the evaluation results of different text detection algorithms in the ICDAR 2011 dataset. In the evaluations of proposed text detection algorithms, our algorithm achieves the best recall and the state-of-art result. The isolate region extract mechanism in I-MSER grantees ensures the extract text regions contain less background units. Thus the text regions can be easily detected which improves the recall. It can be indicated that the I-MSER has advantages in extracting candidate text regions from natural images by achieving the highest recall among the algorithms (0.71 compare to the second best recall 0.66) and the CNN model has advantages in identifying the text regions from the candidate text regions by achieving the precision close to the best (0.82 compare to the best precision 0.83). Table 3 summarizes the evaluation results of different text detection algorithms in the ICDAR 2013 dataset. In the evaluations of proposed text detection algorithms, our algorithm also achieves the best recall and the state-of-art result.

\section{CONCLUSION}

We propose a CNN based text detection algorithm with a modified MSER detector --- I-MSER, which extracts isolated (non-overlapping) regions instead of overlapping regions detected by original MSER from natural images. The I-MSER is motivated by the observation that the statistics of text and background MSERs have significant differences in appearance. I-MSER is designed for extracting isolated object regions which have stable internal color and sharp edges, e.g. texts, blocks. Our I-MSER algorithm has great advantages in detecting text regions addressing challenges such as noise, contrast variations, and plentiful complex backgrounds typical in natural images. We propose a statistical learning method to optimize the main parameter of I-MSER, and the experiment shows that it achieves the best performance in our algorithm. A multi-layer CNN model is trained to identify the candidate text regions extracted by the I-MSER for text detection. We evaluate our algorithm in the ICDAR 2011 and 2015, where our algorithm achieves the best recall and the state-of-art result.

\section*{ACKNOWLEDGEMENT}

The work is partially supported by the National Natural Science Foundation of China(Grant No. 61201384, 61221001) and Shanghai Science and Technology Committees of Scientific Research Project(Grant No. 14DZ110 1200).

\nocite{ICDAR2011,Epshtein2010,Shivakumara2011,
Matas2004,Chen2011,Neumann2012,Tsai2012,Nister2008,ICDAR2013,Huang2014,sermanet2013,
neumann2013,bai2013}
\bibliography{bare_conf}
\bibliographystyle{spiebib} 

\end{document}